\documentclass[a4paper,conference]{IEEEtran}
\IEEEoverridecommandlockouts
\usepackage{cite}
\usepackage{amsmath,amssymb,amsfonts}
\usepackage{algorithmic}
\usepackage{graphicx}
\usepackage{textcomp}
\usepackage{xcolor}
\usepackage{balance}
\usepackage{multirow}
\usepackage{hyperref}
\def\BibTeX{{\rm B\kern-.05em{\sc i\kern-.025em b}\kern-.08em
    T\kern-.1667em\lower.7ex\hbox{E}\kern-.125emX}}
\begin{document}

\title{Electricity Price Forecasting Model based on Gated Recurrent Units }

\author{\IEEEauthorblockN{Nafise Rezaei}
\IEEEauthorblockA{\textit{Faculty of ECE} \\
\textit{Qom University of Technology}\\
Qom, Iran \\
rezaii.n@qut.ac.ir}
\and
\IEEEauthorblockN{Roozbeh Rajabi}
\IEEEauthorblockA{\textit{Faculty of ECE} \\
\textit{Qom University of Technology}\\
Qom, Iran \\
rajabi@qut.ac.ir}
\and
\IEEEauthorblockN{Abouzar Estebsari}
\IEEEauthorblockA{\textit{School of the Built Environment and Architecture} \\
\textit{London South Bank University}\\
London, United Kingdom \\
estebsaa@lsbu.ac.uk}
}

\maketitle

\IEEEpubidadjcol

\begin{abstract}
	The participation of consumers and producers in demand response programs has increased in smart grids, which reduces investment and operation costs of power systems. Also, with the advent of renewable energy sources, the electricity market is becoming more complex and unpredictable. To effectively implement demand response programs, forecasting the future price of electricity is very crucial for producers in the electricity market. Electricity prices are very volatile and change under the influence of various factors such as temperature, wind speed, rainfall, intensity of commercial and daily activities, etc. Therefore, considering the influencing factors as dependent variables can increase the accuracy of the forecast. In this paper, a model for electricity price forecasting is presented based on Gated Recurrent Units. The electrical load consumption is considered as an input variable in this model. Noise in electricity price seriously reduces the efficiency and effectiveness of analysis. Therefore, an adaptive noise reducer is integrated into the model for noise reduction. The SAEs are then used to extract features from the de-noised electricity price. Finally, the de-noised features are fed into the GRU to train predictor. Results on real dataset shows that the proposed methodology can perform effectively in prediction of electricity price.
\end{abstract}


\section{Introduction}

Current technology cannot store electricity in large amounts efficiently and cost effectively. Traditionally, the production has been following the demand to ensure the balance, and in smart grids with integration of more intermittent renewable-sourced production, the demand is also following the generation by appropriate demand side management schemes \cite{RN1}. Electricity demand depends on temperature, wind speed, precipitation, intensity of business and everyday activities, etc \cite{RN2, MDPI_Electronics2020}. These would result in price dynamics. The electricity market actors find it challenging to forecast the electricity load \cite{PowerTech2019}, generation, and price accurately and efficiently as the systems face more uncertainty, non-linearity and volatility. \cite{RN3}. 
There are different methods and solutions to tackle the above-mentioned challenges. However, each method comes with its pros and cons. Empirical Mode Decomposition (EMD) is a technique which decomposes a signal into Intrinsic Mode Functions (IMFs) series and a residual with different frequencies. Besides the advantages of using EMDs, there are two main issues: i) the very extreme values at the two ends of the signal cannot be determined; and ii) the same IMF contains more than two main frequencies, which limits the noise reduction \cite{RN4}. These problems are addressed by Ensemble EMD (EEMD) in \cite{RN5}, where the signals are integrated with white Gaussian noise. However, adding noise can produce varying numbers of IMFs, whose consequence would be observation of residual noise in the reconstructed signals after the decomposition.
The problem of imbalance EMD decomposition scale was resolved when complete EEMD with Adaptive Noise (CEEMDAN)\cite{RN6} added white noise components in the sifting process of each IMF. Selecting the sensitive mode to distinguish relevant and irrelevant IMFs in an efficient way is a serious challenge in EMD-based methods. To tackle these challenges, we use CEEMDAN method which potentially avoid the spurious modes and the reduction in the amount of noise contained in the modes.
In the following sections of this paper, some related works on electricity price forecasting are summarized, an improved version of CEEMDAN as well as a GRU model is introduced, and results on evaluating the efficiency of the new modified method are presented. The paper is concluded then with remarks. 

\section{Literature Review}\label{sec:review}

Creating high quality and effective forecasting models is becoming challenging and difficult in the new electricity markets where most of consumers and producers take active roles in the market, more renewable resources with intermittent behavior are integrated into the network, and more fluctuations and volatility are observed in the system \cite{RN3}. Reviewing literature shows the challenges resulted in developing several different methods and models by involved stakeholders and researchers. These models can be classified into three categories: statistical methods, artificial intelligence methods, and hybrid methods \cite{RN7}. Statistical methods such as autoregressive moving average (ARMA)\cite{RN8}, autoregressive integrated moving average (ARIMA)\cite{RN9}, generalized autoregressive conditional heteroskedasticity (GARCH)\cite{RN10}, vector auto-regression (VAR)\cite{RN11}, and Kalman filters (KF)\cite{RN12}, outperform other methods in stable power markets\cite{RN13}. However, the rapid changes of electricity prices with their nonlinear characteristics are not considered in the electricity price, which is a limitation of these methods \cite{RN3}. Comparing to these statistical methods, artificial intelligence methods, instead, are capable of managing the nonlinear properties and rapid changes\cite{RN14}. There are many studies on the subject of electricity price prediction based on artificial intelligence methods such as artificial neural network (ANN)\cite{RN15,RN16}, recurrent neural network (RNN)\cite{RN17, MDPI_18_RNN}, and Extreme Learning Machine (ELM) \cite{RN18}. Nevertheless, applying only traditional models may overlook the complex characteristics of the original nonlinear and non-fixed electricity prices. To overcome the disadvantages of single models, many hybrid models have been developed for electricity price forecasting that use data analysis methods to preprocess nonlinear and nonstationary electricity price data before forecasting\cite{RN14}. For example, Zhang et al\cite{RN19}, developed a forecasting model using WT models, generalized regression neural network (GRNN) and GARCH models for electricity price forecasting, which was validated in the Spanish electricity market and performed well \cite{RN20}.

\section{Proposed Method}\label{sec:proposed}
\begin{figure}[htbp]
	\centerline{\includegraphics[width=8cm]{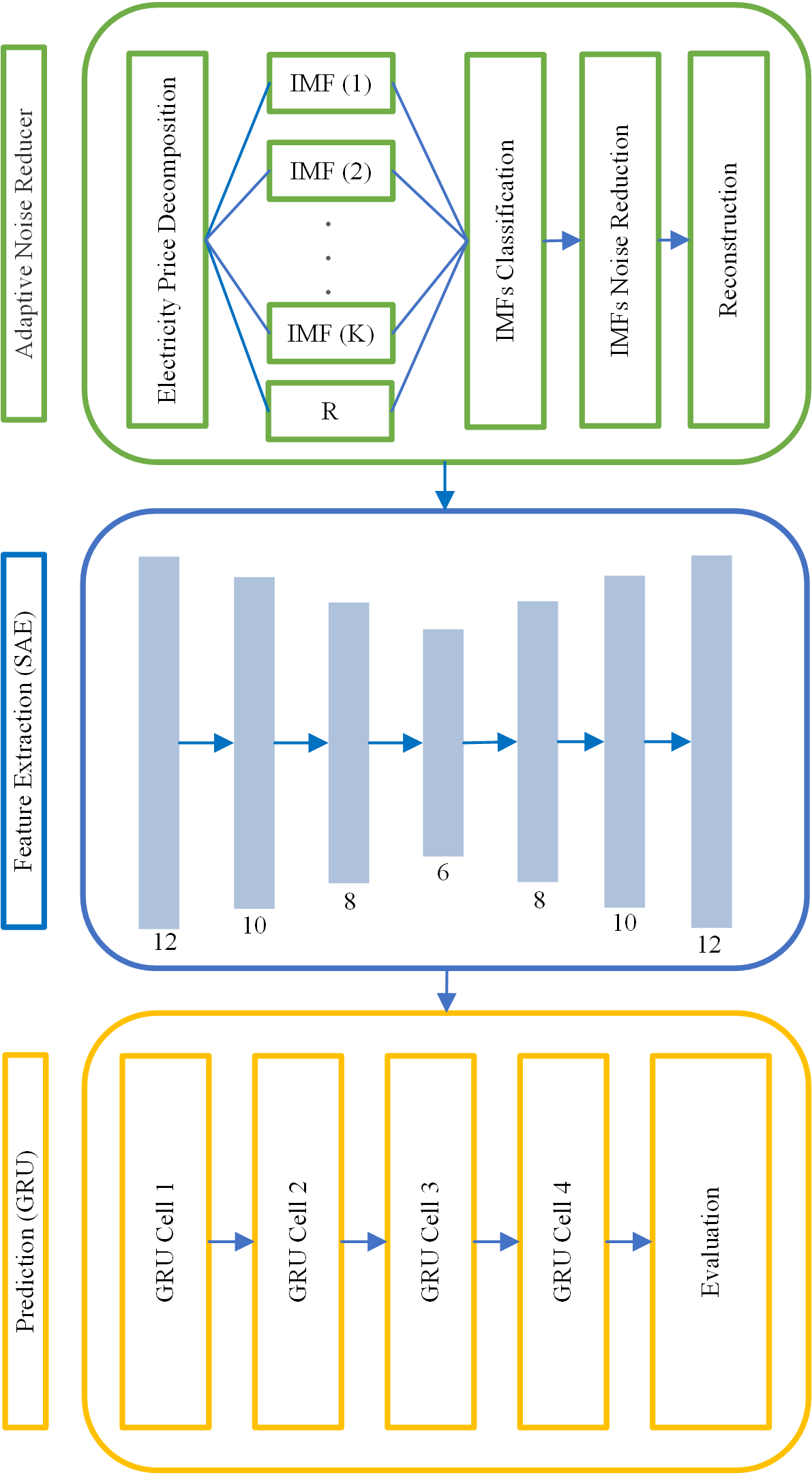}}
	\caption{The proposed ANR-SAE-GRU method.}
	\label{fig:Proposed}
\end{figure}

Increasing attention to the prediction of time series such as Electricity Price indicates its potential application in many fields. However, predicting  time series is not an easy task. How to deal with noise and extract features from  time series are the two main challenges to obtain accurate predictions. Because time series are unstable and often contain noise, data processing is necessary to reduce the effect of noise on prediction. To solve these issues, a framework for noise removing and feature extraction for electricity price forecasting is proposed. In this framework, CEEMDAN technique is used to decompose time series and eliminate noise. A combination of stacked encoders and GRU is also used for feature extraction and prediction. The diagram of the proposed model is presented in Figure \ref{fig:Proposed}.

\subsection{Adaptive Noise Reduction}
First, CEEMDAN technique \cite{RN6} is used to decompose electricity price into IMFs and a residual (R). This technique can decompose a signal into a set of IMFs that include noise modes and information modes. Therefore, it can be a powerful compatible tool for extracting IMFs for signal reconstruction. The main issue is how to choose a sensitive mode to distinguish between IMFs and unrelated IMFs in an efficient way.   residual has the lowest frequency and the frequency distribution of IMFs is from top to bottom. It is assumed that the electricity price time series are  $\mathrm{X}=\left\{\mathrm{x}_{1}, \mathrm{x}_{2}, \ldots, \mathrm{x}_{\mathrm{m}}\right\}$ \cite{Ensemble_Access19}.

Step 1: The amount of Gaussian noise $\epsilon_0w^i(i=1,2,\cdots,I)$ is added to the electricity price and $X+\epsilon_0w^i$ is decomposed by the EMD technique to obtain the first IMF:

\begin{equation}
	\begin{aligned}
	&IMF_1=\frac{1}{I}\sum_{i=1}^{I}E_1(X+\epsilon_0w^i)
	\end{aligned}
\end{equation}

In this relation, $\epsilon_0$ is the adaptive coefficient and $I$ is the number of adding Gaussian noise. In this paper, the value of $I$ is considered equal to 100.

Step 2: When $k=1,2,\cdots,I$ the remaining $k$th residue is calculated as $r_k=r_{k-1}-IMF_k$ and $r_0=X$. Then $r_k+\epsilon_0E_k(w^i)(i=1,2,\cdots,I)$ is decomposed by EMD to obtain the first state of EMD and the definition of the $(k+1)$th of the IMF is as follows:

\begin{equation}
	\begin{aligned}
		&IMF_{k+1}=\frac{1}{I}\sum_{i=1}^{I}E_1(r_k+\epsilon_kE_k(w^i))
	\end{aligned}
\end{equation}

Step 3: Step 2 are repeated until the residue cannot be decomposed anymore. The final residue is as follows:

\begin{equation}
	\begin{aligned}
		&R=X-\sum_{k=1}^{K}IMF_k
	\end{aligned}
\end{equation}

Where $K$ is the IMF number and the electricity price time series can be rewritten as follows:
\begin{equation}
	\begin{aligned}
		&X=\sum_{k=1}^{K}IMF_k+R
	\end{aligned}
\end{equation}

In the next step, noisy and no-noise IMF must be separated. To do this, the permutation entropy method is used. Permutation Entropy (PE) \cite{PE_PhysRevLett} is a method of calculating the complexity of time series. The permutation entropy measures the randomness of time series. The higher the permutation entropy, the more noise will present in time series. IMF classification is based on the PE value of each IMF. The PE value threshold is set at 0.7 based on several experiments in different methods. If the PE value of an IMF is higher than the threshold, the IMF contains noise; otherwise, the IMF is noise-free. Assuming that P is the point of separation between high-frequency noise IMFs and low-frequency noise-free IMFs, the decomposed series can be expressed as follows:

\begin{equation}
	\begin{aligned}
		&X = \sum_{k=1}^{P-1}IMF^{(k)} + \sum_{k=p}^{K}IMF^{(k)}+R
	\end{aligned}
\end{equation}

\subsection{Noise Reduction}

To reduce noise in the IMF with high-frequency noise, the adaptive threshold  and the threshold function are determined using the following equations:

\begin{equation}
	\begin{aligned}
		&\lambda=\sigma\sqrt{\frac{2ln(m)}{ln(k+1)}}\\
		&w_\lambda=\begin{cases}
			sgn(w)(|w|-\lambda), & |w| \ge \lambda \\
			0, & |w| < \lambda
		\end{cases}
	\end{aligned}
\end{equation}

where $\sigma$, $m$ and $K$ are the standard deviation, length and number of IMFs, respectively. The notation $w$ is a noise value in the IMF. If the absolute value of $w$ is greater than $\lambda$, it takes the value $sgn(w)(|w|-\lambda)$, otherwise it takes 0 \cite{Ensemble_Access19}.

\subsection{Reconstruction}

Noise-free time series are obtained by adding high-frequency and low-frequency noise-free IMFs. Noise-free time series are obtained as follows:

\begin{equation}
	\begin{aligned}
		&X = \sum_{k=1}^{P-1}IMF^{(k)} + \sum_{k=p}^{K}IMF'^{(k)}+R
	\end{aligned}
\end{equation}

Where $IMF'^{(k)}(k=P,P+1,\cdots,K)$ is equivalent to the high-frequency noise-free IMF and  $IMF^{(k)}$ is equivalent to low frequency noise-free IMF and R is the residue.

\subsection{Unsupervised Learning}

\begin{figure}[htbp]
	\centerline{\includegraphics[width=7cm]{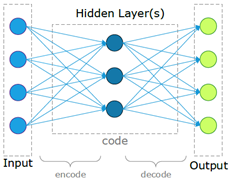}}
	\caption{Auto Encoder Structure.}
	\label{fig:SAE}
\end{figure}

To effectively analyze the price of electricity, an important method is to extract the features and their interdependence between variables. Feature extraction methods can be divided into supervised learning methods and unsupervised learning methods. Supervised learning method  extracts functions from labeled datasets. However, electricity price are often unlabeled and converting it to labeled data is costly and requires professional knowledge.  unsupervised learning,  can learn a feature display layer from unlabeled data. In addition, these layers can be stacked to create deep grids. The proposed model uses automatic stack encoder (SAE), which is a kind of unsupervised learning method. Figure \ref{fig:SAE} shows the structure of a typical encoder. SAE consist of two parts, the encoder and the decoder. In this paper, an auto encoder with three hidden layers is used, which extracts the required features to increase the accuracy of forecasting.

\subsection{GRU Predictor}
Gated Recurrent Unit (GRU) is especially effective in learning the characteristics of longer time series, because learning the properties of long sequences requires less training weights and faster calculations compared to LSTM. The GRU is therefore used here to process and learn the time domain characteristics of the electricity prices extracted by SAE and to improve the accuracy of the final model forecast.

\section{Evaluation and Results}\label{sec:experiments}

In this section, the proposed framework is compared with the framework, which used LSTM to predict electricity price. First, the dataset is described, and then the method for data sampling for evaluation is described. Three criteria are used to evaluate the prediction models, including root mean square error (RMSE) and mean absolute error (MAE).

\subsection{Dataset}
The data used in this article are related to the price of electricity consumed in Iran over three years. In this dataset, hourly electricity prices are collected. In total, the size of the electricity price data set is 35040. In this study, information for 1186 days is used for training (28032 data samples, one sample per hour) and 292 days for testing (7008 samples). Iran electricity market price dataset is public and can be downloaded from \href{https://www.igmc.ir/Electronic-Services/Power-Market-Deputy/Reports}{https://www.igmc.ir/Electronic-Services/Power-Market-Deputy/Reports}.

\begin{table*}[htbp]
	\caption{Comparison Results Based on RMSE and MAE.}
	\begin{center}
		\begin{tabular}{|c|c|c|c|c|c|c|c|c|c|}
			\hline
			\multirow{2}{*}{Model}&{Measure}&\multicolumn{4}{|c|}{RMSE}&\multicolumn{4}{|c|}{MAE}\\\cline{2-10}
			&{Horizon (Hour)}&{3}&{6}&{9}&{12}&{3}&{6}&{9}&{12} \\
			\hline
			\multicolumn{2}{|c|}{\textbf{Proposed Method}} &{2.86}&{3.89}&{5.18}&{6.48}&{1.8}&{2.83}&{3.92}&{5}\\
			\hline
			\multicolumn{2}{|c|}{\textbf{ANR-SAE-LSTM}} & {4.33}& {5.28}& {6.39}& {7.5}& {2.09} & {2.94} & {4}& {5.42}\\
			\hline
		\end{tabular}
		\label{tbl:comparison}
	\end{center}
\end{table*}

\subsection{Evaluation Measures}
The evaluation process was performed using evaluation indicators named RMSE and MAE for both training and test sets. Eqs. \ref{eq:RMSE}, \ref{eq:MAE} show these criteria, respectively. 

 \begin{equation}
 	\mbox{RMSE}=\sqrt{\frac{1}{N}\sum_{i=1}^{N}(y_i-\hat{y_i})^2}
 	\label{eq:RMSE}
 \end{equation}
 
 \begin{equation}
 	\mbox{MAE}=\frac{1}{N}\sum_{i=1}^{N}|y_i-\hat{y_i}|
 	\label{eq:MAE}
 \end{equation}
 where $N$ is the total number of sample points. $y_i$ and $\hat{y_i}$ are the actual and predicted values of $i$th sample respectively.

\subsection{Results and discussion}\label{subsec:dataset}
In this section, the proposed GRU-based method is compared with the LSTM-based method on a electricity price dataset to demonstrate the superiority of the proposed method in  electricity price prediction. Simulations are performed in Python using NumPy, PyEEMD, and Keras libraries.

In our experiment,  the prediction horizon  is set to {3, 6, 9, 12} and the corresponding input data length  is set to 24. Table \ref{tbl:comparison} summarizes the prediction results according to different values of the forecast horizon. It can be seen that as the forecast horizon increases, the forecasting error for both methods becomes larger, but the proposed method works better for all three criteria and has the least error growth. On average, for different forecast horizons, the proposed method is 1.54 and 0.3 superior to the other method in terms of RMSE and MAE criteria, respectively.

\begin{figure}[htbp]
	\centerline{\includegraphics[width=9cm]{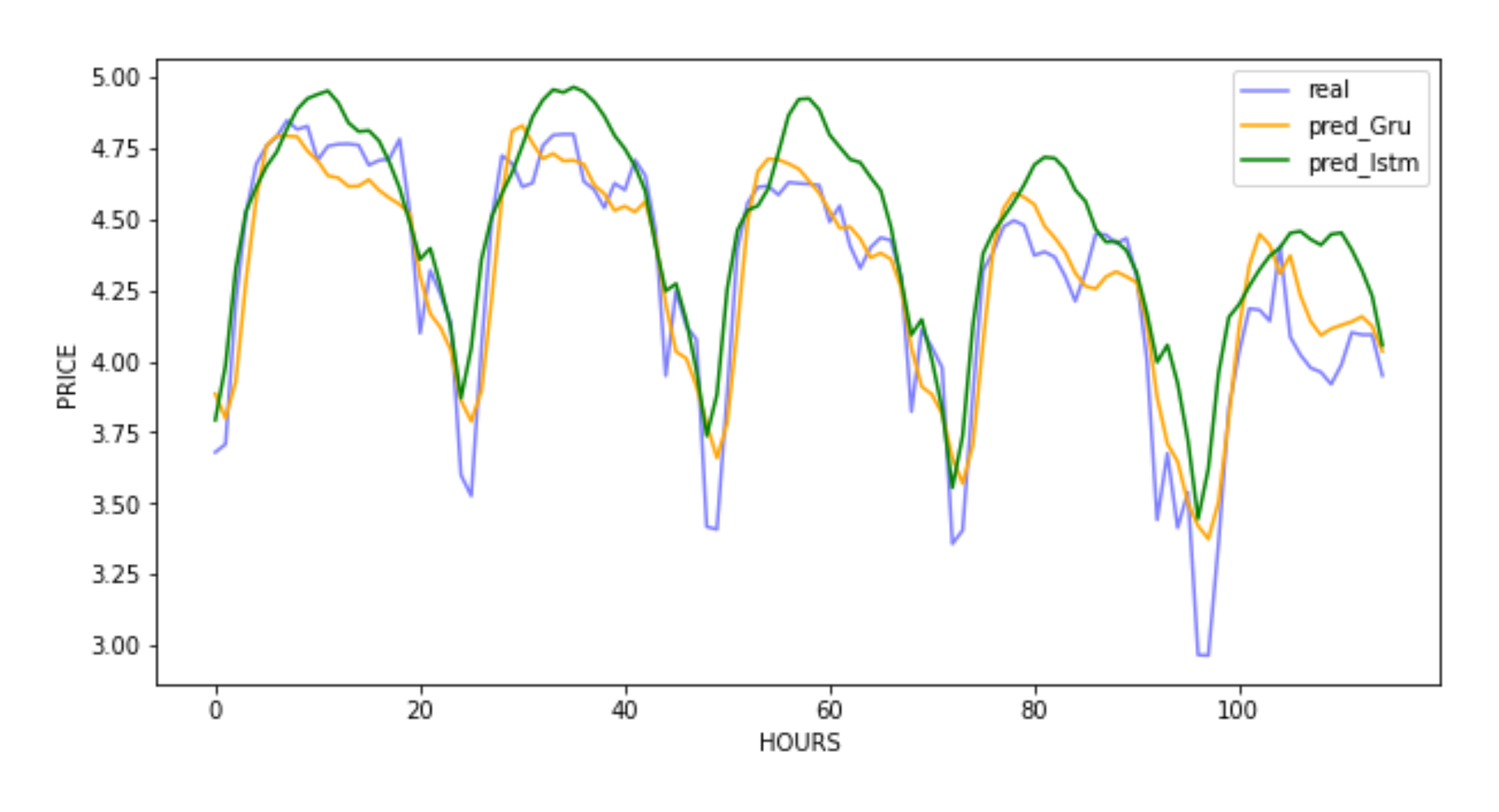}}
	\caption{Prediction results compared to actual values.}
	\label{fig:Result}
\end{figure}

Figure \ref{fig:Result} shows the forecast results of the methods compared to the actual cost values for 12 days. We can see that the proposed method provides the predicted values closer to the actual values than the ANR-SAE-LSTM.

\section{Conclusion}\label{sec:conclusion}
To deal with noise in electricity price, an ANR method was used in this paper. Permutation Entropy (PE) is used to determine the boundary point between noisy and low-noise IMFs, and an adaptive threshold function is constructed to eliminate noise from high-frequency IMFs. SAE was used to extract features that fully take into account  electricity price forecasting performance and prevent overfitting, the GRU model was employed to obtain a robust forecaster. The proposed GRU-based method was compared with the LSTM-based method on a power consumption price dataset. Two criteria of RMSE and MAE were considered for the results comparsion. The results showed that the forecast error for both methods becomes larger as the forecast horizon increases. On average, for different forecast horizons, the proposed method is 1.54 and 0.3 superior to the other method in terms of RMSE and MAE criteria, respectively. 

Our approach is only to forecast electricity prices using historical electricity price data. However, since the price of electricity depends on several factors, the effect of several dependent variables on the price forecast can be considered. In future work, we will further explore the possible methods of multi-objective forecasting. In the proposed method, only one GRU predictor is used in the aggregation stage, while several predictors such as LSTM and GRU can combine consumption patterns in a group.

\balance

\bibliographystyle{IEEEtran}
\bibliography{IEEEabrv,refs}

\end{document}